\documentclass[sigconf]{acmart}
\renewcommand\footnotetextcopyrightpermission[1]{}
\usepackage{xcolor}
\usepackage{colortbl}
\usepackage{multirow}
\usepackage{makecell}
\usepackage{pifont}
\usepackage{cuted}
\usepackage{cleveref}
\usepackage{xspace}
\newcommand{\ie}{\textit{i.\,e.}\xspace}
\newcommand{\eg}{\textit{e.\,g.}\xspace}

\AtBeginDocument{%
  }

\begin{document}

\title{BiCoord: A Bimanual Manipulation Benchmark towards Long-Horizon Spatial-Temporal Coordination}

\author{Xingyu Peng}
\authornote{Equal Contribution.}
\email{pengxyai@buaa.edu.cn}
\affiliation{%
  \institution{Beihang University}
  \institution{Zhongguancun Academy}
  \country{}
}

\author{Chen Gao}
\authornotemark[1]
\email{gaochen.ai@gmail.com}
\affiliation{%
  \institution{Beihang University}
  \institution{National University of Singapore}
  \country{}
}

\author{Liankai Jin}
\authornotemark[1]
\email{liankaijin@buaa.edu.cn}
\affiliation{%
  \institution{Beihang University}
  \country{}
}

\author{Annan Li}
\email{lianna@buaa.edu.cn}
\affiliation{%
  \institution{Beihang University}
  \country{}
}

\author{Si Liu}
\authornote{Corresponding Author.}
\email{liusi@buaa.edu.cn}
\affiliation{%
  \institution{Beihang University}
  \country{}
}

\begin{abstract}
Bimanual manipulation, i.e., the coordinated use of two robotic arms to complete tasks, is essential for achieving human-level dexterity in robotics. Recent simulation benchmarks, e.g., RoboTwin and RLBench2, have advanced data-driven learning for bimanual manipulation. However, existing tasks are short-horizon and only loosely coordinated, failing to capture the spatial-temporal coupling inherent in real-world bimanual behaviors. To address this gap, we introduce BiCoord, a benchmark for long-horizon and tightly coordinated bimanual manipulation. Specifically, BiCoord comprises diverse tasks that require continuous inter-arm dependency and dynamic role exchange across multiple sub-goals. Also, we propose a suite of quantitative metrics that evaluate coordination from temporal, spatial, and spatial-temporal perspectives, enabling systematic measurement of bimanual cooperation. Experimental results show that representative manipulation policies, e.g., DP, RDT, Pi0, and OpenVLA-OFT, struggle with long-duration and highly coupled tasks, revealing fundamental challenges in achieving long-horizon and tight coordination tasks. We hope BiCoord can serve as a foundation for studying long-horizon cooperative manipulation and inspire future research on coordination-aware robotic learning. All datasets, codes and supplements could be found at \url{https://buaa-colalab.github.io/BiCoord/}.
\end{abstract}

\begin{teaserfigure}
  \includegraphics[width=\textwidth]{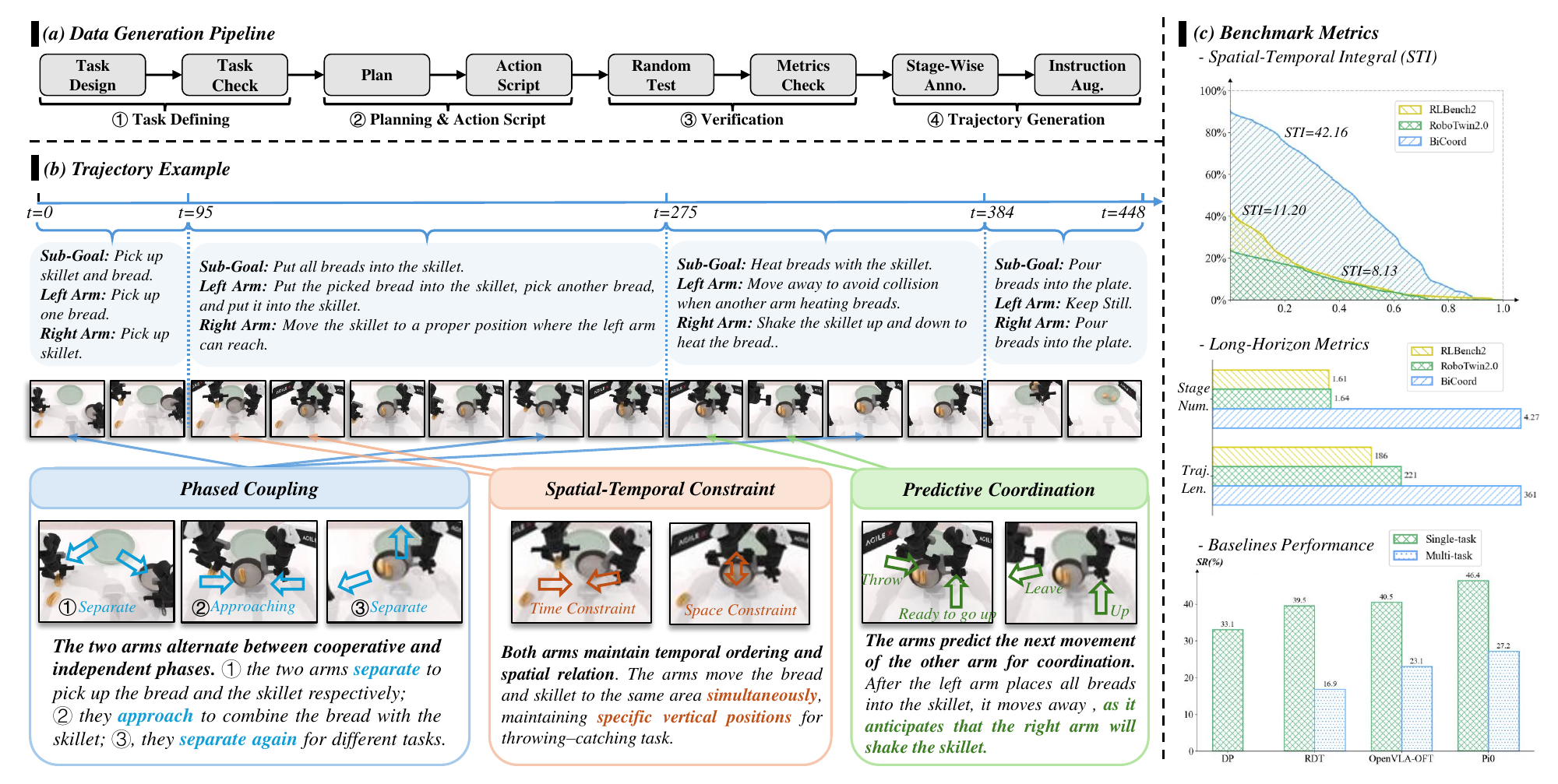}
  \caption{\emph{\textbf{
Overview of BiCoord}}. (a) The data generation pipeline. (b) An example trajectory of \textit{Cook} task is exhibited. Each trajectory is divided into several stages with sub-goals and arm behaviours. Besides, key features of bimanual coordination are embodied in BiCoord, like phased coupling, spatial-temporal constraint and predictive coordination. (c) We design metrics to evaluate the bimanual manipulation benchmarks. STI characterizes the temporal and spatial coupling of the dual arms, and long-horizon metrics reflect task length. We also evaluate four methods on BiCoord in both single-task and multi-task setting.}
  \Description{Enjoying the baseball game from the third-base
  seats. Ichiro Suzuki preparing to bat.}
  \label{fig:teaser}
\end{teaserfigure}

\begin{abstract}
Bimanual manipulation, i.e., the coordinated use of two robotic arms to complete tasks, is essential for achieving human-level dexterity in robotics. Recent simulation benchmarks, e.g., RoboTwin and RLBench2, have advanced data-driven learning for bimanual manipulation. However, existing tasks are short-horizon and only loosely coordinated, failing to capture the spatial-temporal coupling inherent in real-world bimanual behaviors. To address this gap, we introduce BiCoord, a benchmark for long-horizon and tightly coordinated bimanual manipulation. Specifically, BiCoord comprises diverse tasks that require continuous inter-arm dependency and dynamic role exchange across multiple sub-goals. Also, we propose a suite of quantitative metrics that evaluate coordination from temporal, spatial, and spatial-temporal perspectives, enabling systematic measurement of bimanual cooperation. Experimental results show that representative manipulation policies, e.g., DP, RDT, Pi0, and OpenVLA-OFT, struggle with long-duration and highly coupled tasks, revealing fundamental challenges in achieving long-horizon and tight coordination tasks. We hope BiCoord can serve as a foundation for studying long-horizon cooperative manipulation and inspire future research on coordination-aware robotic learning. All datasets, codes and supplements could be found at \url{https://buaa-colalab.github.io/BiCoord/}.
\end{abstract}
\maketitle

\section{Introduction}
\label{sec:intro}

Bimanual manipulation, \ie, the coordinated use of two robotic arms to interact with the environment for completing manipulation tasks, is a fundamental yet challenging capability in robotics. It enables robots to perform complicated tasks that require dexterous coordination.

Compared with single-arm, dual-arm/bimanual manipulation offers significant advantages. This is deeply rooted in the way our physical world is designed, \ie, tools, household objects, and most human-centric environments are naturally made for two-handed use. Human bimanual actions (\eg, cooking) exhibit strong inter-arm dependency, reflecting a high degree of spatial and temporal coordination. The two arms cooperate in a spatially complementary manner (\eg, stabilizing and operating simultaneously) and synchronize temporally to achieve fluid, goal-oriented motion. Such spatially and temporally entangled bimanual interactions are crucial for achieving human-level dexterity, motivating the design of bimanual manipulation that can mirror this intrinsic coordination.

Recently, the community recognizes the importance of bimanual manipulation, and simulation-based benchmarks are developed, such as RoboTwin~\cite{mu2025robotwin,chen2025robotwin} and RLBench2~\cite{grotz2024peract2}. These benchmarks provide bimanual manipulation tasks and expert demonstration data for facilitating data-driven learning. 
However, existing benchmarks still fall short in two aspects for capturing the full complexity of real-world coordination. 
\emph{\textbf{(1) Short-horizon tasks:}} Existing benchmarks only focus on short tasks that can be completed within a few motion primitives (\eg, picking and placing). Such tasks fail to reflect the long-horizon dependencies and hierarchical structure that naturally arise in real-world manipulation, where two arms must continuously coordinate across multiple sub-goals, object states, and contact transitions.
\emph{\textbf{(2) Loosely coordinated bimanual tasks:}} While these benchmarks include two-arm scenarios, the majority of them involve only weak coordination between arms, where each arm operates largely independently or performs temporally disjoint actions. In contrast, human-like manipulation requires tight spatial-temporal coupling, such as stabilizing and manipulating an object simultaneously, or dynamically exchanging forces and roles between arms. The lack of such strongly coordinated bimanual tasks hinders the development of models capable of understanding and executing truly cooperative manipulation behaviors.

In this work, we propose \emph{\textbf{BiCoord}}, a bimanual manipulation benchmark towards long-horizon and closely coordinated tasks. As shown in \cref{fig:teaser}(c), BiCoord exhibits significantly longer trajectories and more stages compared to existing benchmarks. 
Moreover, our tasks are designed to capture tight coordination, as exemplified by the \textit{Cook} task shown in \cref{fig:teaser}(b). 
Specifically, the coordination between the two arms manifests three characteristics.
\emph{\ding{182} Phased coupling:} the two arms alternate between cooperative and independent phases, reflecting task-dependent synchronization and decoupling. \eg, the two arms alternately separate and approach to complete pick-up and throwing-catching tasks while avoiding collisions.
\emph{\ding{183} Spatial-temporal constraints:} both arms must maintain temporal ordering and spatial relationships to complete the task. For example, the two arms reach the same spatial region at the same time, and maintain vertical positions to complete the throwing–catching task.
\emph{\ding{184} Predictive coordination:} each arm anticipates the future motion or state of the other arm, enabling proactive planning of its own trajectory. For instance, the left arm proactively retreats to prevent collision, anticipating the right arm's upcoming shaking motion.

We further design a set of metrics to quantitatively assess bimanual coordination from multiple views, \eg, spatial, temporal and spatial-temporal perspectives. As shown in \cref{fig:teaser}(c) and \cref{tab:comparison}, metrics indicate that our BiCoord features long-horizon tasks with tight spatial-temporal bimanual coordination. For instance, the Spatial-Temporal Integral (STI) measures the degree of spatial-temporal coupling, and BiCoord achieves substantially higher STI value than existing benchmarks.
Based on BiCoord, we systematically conduct extensive experiments by training and evaluating representative learning-based bimanual manipulation methods, including DP~\cite{chi2023diffusion}, RDT~\cite{liu2024rdt1b}, Pi0~\cite{black2024pi_0} and OpenVLA-OFT~\cite{openvla_oft}. Both quantitative and qualitative analyses yield a series of findings and insights, shedding light on the challenges and characteristics of tightly coordinated bimanual manipulation. All data, code and checkpoints will be released to facilitate related research.
\section{Related Works}
\label{sec:rel}

\begin{table*}[!htpb]
\centering
\caption{\textbf{Comparison of bimanual manipulation benchmarks.} All metrics are averaged on trajectories within the dataset. SMP, MAD, AAD, STI are presented in percentage (\%). TL reflects the timesteps that required to complete a task. SN is the number of stages in a task. ON shows the number of objects that need to be manipulated or focused in a task. 
}
\vspace{-2mm}
\resizebox{\textwidth}{!}{
{
\renewcommand{\arraystretch}{1.35}
\setlength{\tabcolsep}{5pt}
\begin{tabular}{r||c||c||ccccc||ccc}
\hline

 \rowcolor[HTML]{f8f9fa} \multicolumn{1}{c||}{}& \multicolumn{1}{c||}{} & \multicolumn{1}{c||}{} & \multicolumn{5}{c||}{Spatial-Temporal Metrics} & \multicolumn{3}{c}{Long-Horizon Metrics} \\
\rowcolor[HTML]{f8f9fa} \multicolumn{1}{c||}{\multirow{-2}{*}{Benchmark}} & 
\multicolumn{1}{c||}{\multirow{-2}{*}{\makecell[c]{Task \\ Number}}}  & 
\multicolumn{1}{c||}{\multirow{-2}{*}{\makecell[c]{Stage-Wise \\ Annotation \\ \& Evaluation}}} & 
\makecell[c]{Simultaneous \\ Movement \\ Time (SMT) $\uparrow$} & 
\makecell[c]{Simultaneous \\ Movement \\ Percentage (SMP,\%) $\uparrow$} & 
\makecell[c]{Minimum \\ Relative \\ Distance (MRD,\%) $\downarrow$} & 
\makecell[c]{Average \\ Relative \\ Distance (ARD,\%) $\downarrow$} & 
\makecell[c]{Spatial-Temporal \\ Integral (STI,\%) $\uparrow$} & 
\makecell[c]{Trajectory \\ Length (TL) $\uparrow$} & 
\makecell[c]{Stage \\ Number (SN) $\uparrow$} & 
\makecell[c]{Object \\ Number (ON) $\uparrow$} \\
\hline
\hline

RLBench2 & 13 & \ding{55} & 179 & \textbf{97.10} & 54.57 & 114.93 & 11.20 & 186 & 1.61 & 2.23 \\
RoboTwin2.0 & 50 & \ding{55} & 60 & 26.10 & 63.83 & 82.37 & 8.13 & 221 & 1.64 & 2.02 \\
BiCoord (Ours) & 18 & \ding{51} & \textbf{329} & 92.81 & \textbf{29.59} &\textbf{ 55.77} & \textbf{42.16} & \textbf{361} & \textbf{4.27} & \textbf{3.66} \\

\hline
\end{tabular}
}
}
\vspace{-1mm}
\label{tab:comparison}
\end{table*}
\noindent\textbf{Simulators and Benchmarks for Robotic Manipulation.}
Recently, numerous simulators and benchmarks have been proposed to boost the research of robot manipulation. For simulators~\cite{xiang2020sapien,robocasa2024,taomaniskill3,robosuite2020,mittal2023orbit,todorov2012mujoco,makoviychuk2021isaac}, Mujoco~\cite{todorov2012mujoco}, SAPIEN~\cite{xiang2020sapien} and Isaac Sim~\cite{makoviychuk2021isaac} are widely used in manipulation area. 
For datasets and benchmarks, preliminary works mainly focuses on unimanual manipulation~\cite{metaworld,mees2022calvin,liu2023libero,ebert2021bridge,walke2023bridgedata,han2025robocerebra,james2020rlbench,khazatsky2024droid,wang2022bulletarm,dasari2021rb2}. 
By designing tasks varying on objects, layouts and goals, LIBERO~\cite{liu2023libero} effectively evaluates the policy from multiple perspectives. 
 RLBench~\cite{james2020rlbench} provides convenient tools for task customization. 
RoboCerebra~\cite{han2025robocerebra} provides a benchmark towards long-horizon manipulation. 
Some works also pay attention to building a unified platform for robot manipulation with diverse embodiments~\cite{geng2025roboverse,bu2025agibot,wu2024robomind,openxembodiedment,lan2025autobio,li2025labutopia,sferrazza2024humanoidbench,TinyVLA,dexhanddiff}.
Recently, more datasets and benchmarks of bimanual manipulation emerge~\cite{mu2025robotwin,chen2025robotwin,grotz2024peract2,fan2023arctic,chatzilygeroudis2020benchmark}. RLBench2~\cite{grotz2024peract2} extends the original RLBench~\cite{james2020rlbench} to bimanual manipulation. RoboTwin~\cite{mu2025robotwin,chen2025robotwin} builds a bimanual manipulation platform, allowing efficient training of algorithms and custom manipulation tasks. 
However, current benchmarks for bimanual manipulation primarily focus on short and simple tasks, thus cannot fully reflect models' capabilities in long-horizon reasoning and bimanual coordination. To address this gap, we propose BiCoord, a benchmark designed for long-horizon bimanual manipulation and spatial-temporal coordination.

\noindent\textbf{Learning Methods for Robot Manipulation.} 
Primary research focuses on handling a specific manipulation task~\cite{chen2025g3flow,chi2023diffusion,ze20243d,ke20243d,adaptdiffuser,liang2024skilldiffuser,wang2024rise,wen2025dexvla}. Recently, multi-task manipulation is feasible by VLAs~\cite{brohan2022rt-1, brohan2023rt-2,liu2024rdt1b,black2024pi_0,openvla,openvla_oft,li2024cogact,team2024octo,lapa}. RDT~\cite{liu2024rdt1b} unifies the action representations of different robots. Pi0~\cite{black2024pi_0} combines a flow matching head with a pre-trained vision-language model to inherit Internet-scale semantic knowledge. OpenVLA-OFT~\cite{openvla_oft} displaces the diffusion head in VLAs with a parallel decoder, improving the efficiency of action generation. 
Meanwhile, some works are focused on bimanual manipulation~\cite{smith2012dual,zakka2023robopianist,liu2022robot,zhao2022dualafford,grannen2023stabilize,li2023efficient,aldaco2024aloha,zhao2023learning,fu2024mobile,grotz2024peract2,lu2025anybimanual,gao2024bi,liu2025voxact,jiang2025rethinking,lv2025spatial,yang2025gripper,liu2025d,yu2025manigaussian++,hao2026abstracting}. AnyBimanual~\cite{lu2025anybimanual} extracts action templates from unimanual data, which are synthesized for bimanual actions. DIF~\cite{jiang2025rethinking} achieves coordination through parameter-based communication. KStar Diffuser~\cite{lv2025spatial} reduces the conflicts with knowledge from kinematics and robot structures. PPI~\cite{yang2025gripper} considers the trade-off between spatial awareness and movement continuity. Voxact-b~\cite{liu2025voxact} and ManiGaussian++~\cite{yu2025manigaussian++} adopt the leader-follower architecture. 
However, by baselining mainstream policies on BiCoord, we find the dilemmas of these policies when dealing with long-duration and highly collaborative bimanual tasks, providing insights for related research.

\section{Quantifying Bimanual Coordination}
\label{sec:met}

\subsection{Preliminaries}
Bimanual manipulation requires dual arms to complete a task based on instruction, multi-view observation and robot state. In our setting, besides the head-view camera, two wrist cameras are attached to the left arm and the right arm respectively. At each timestep $t$, three RGB images $I_t = [I_t^{\text{head}},I_t^{\text{left}},I_t^{\text{right}}]$ are captured by these cameras. With instruction $T$ and robot state $S_t$, the actions of the next $H$ timesteps are predicted by the policy $\pi$:
\begin{equation}
    [A_t,\cdots,A_{t+H-1}] = \pi(T,I_t,S_t).
\end{equation}

As for robot state, there are mainly two representation methods in the manipulation area. Some policies denote the robot state as the angles of all joints, while others use end-effector pose and gripper state for representation. For convenience of discussion, we only use the latter in this paper, yet both representations are supported in our benchmark. Specifically, the robot state $S_t$ could be defined as:
\begin{equation}
    S_t=[E_t^{\text{left}},G_t^{\text{left}},E_t^{\text{right}},G_t^{\text{right}}],
\end{equation}
where $E_t$ is the end-effector pose, $G_t\in [0,1]$ is the opening size of the gripper. The end-effector pose is notated by $E_t=(p_t,q_t)$, where $p_t\in \mathbb{R}^3$ is the 3D coordination, $q_t\in \mathbb{H}$ is a quaternion representing orientation. The action $A_t$ could be represented in a similar way, which indicates the desired robot state at timestep $t$.

\subsection{Spatial-Temporal Coordination Metrics}

\label{sec:met:met}

The ways of bimanual collaboration vary from task to task in previous benchmarks, yet no metrics could describe these behaviours from a bimanual coordination perspective. 
Thus, we propose a series of metrics to statistically investigate from the perspective of spatial-temporal coordination.

Basically, dual arms share both space and time during the manipulation process. For space, tasks that require the dual arms to collaborate at a short distance are generally harder than those not. For example, ``\textit{threading a needle}'' is harder than ``\textit{pressing two different buttons}''. Inspired by this, we propose \textbf{Minimum Relative Distance (MRD)} and \textbf{Average Relative Distance
(ARD)}:
\begin{equation}
    MRD = \min_{1\leq t\leq L}\frac{||p_t^{\text{left}}-p_t^{\text{right}}||_2}{||p_1^{\text{left}}-p_1^{\text{right}}||_2},
\end{equation}
\begin{equation}
    ARD = \frac{1}{L}\sum_{t=1}^L\frac{||p_t^{\text{left}}-p_t^{\text{right}}||_2}{||p_1^{\text{left}}-p_1^{\text{right}}||_2},
\end{equation}
where $p_t$ is the end-effector coordination at timestep $t$, $L$ is trajectory length. Note that relative distance is used here to avoid different distance measurements across benchmarks. Generally, smaller MRD and ARD indicate that dual arms need to collaborate at a shorter distance. 

For time, taking actions in parallel is more efficient yet harder for dual arms. For example, ``\textit{putting away plates and cups simultaneously}'' is faster yet harder than ``\textit{putting away plates and cups one by one}''. We define that an arm is taking an action if it is moving or grasping something:
\begin{equation}
    m_t=\begin{cases}
    1,& p_t\not = p_{t-1} \lor G_t \text{ is close}\\
    0,& \text{else}
    \end{cases}
\end{equation}
where $m_t=1$ means the arm is taking an action, $G_t$ is gripper state. Then \textbf{Simultaneous Movement Time (SMT)} and \textbf{Simultaneous Movement Percentage (SMP)} are defined:
\begin{equation}
    SMT = \sum_{t=1}^L m_t^{\text{left}}\cdot m_t^{\text{right}},
\end{equation}
\begin{equation}
    SMP = \frac{SMT}{L}.
\end{equation}
SMT can intuitively display the time of simultaneous action, while SMP is independent of trajectory length, which is convenient to compare across different tasks.

While above metrics measure a task either from space (collaborate at short distance) or from time (take action simultaneously), space and time are also tightly coupled in bimanual coordination. To describe this, we propose \textbf{Spatial-Temporal Integral (STI)}. Specifically, we consider SMP at different arm distance threshold $d$:
\begin{equation}
    SMP_{<d}=\frac{1}{L}\sum_{t=1}^L m_t^{\text{left}}\cdot m_t^{\text{right}}\cdot\chi(\frac{||p_t^{\text{left}}-p_t^{\text{right}}||_2}{||p_1^{\text{left}}-p_1^{\text{right}}||_2}<d),
\end{equation}
where $\chi(P)=1$ if proposition $P$ is true, otherwise $\chi(P)=0$. Then STI is obtained via:
\begin{equation}
    STI = \int_0^1 SMP_{<1-d}\text{d}d.
\end{equation}
By quantitatively describing the spatio-temporal couple of dual arms, STI provides a new perspective to measure the coordination degree of bimanual manipulation tasks.

\subsection{Comparison of Existing Benchmarks}
\label{sec:met:eval}

\begin{figure}[t]
  \centering 
  \resizebox{1\linewidth}{!}{
  \includegraphics{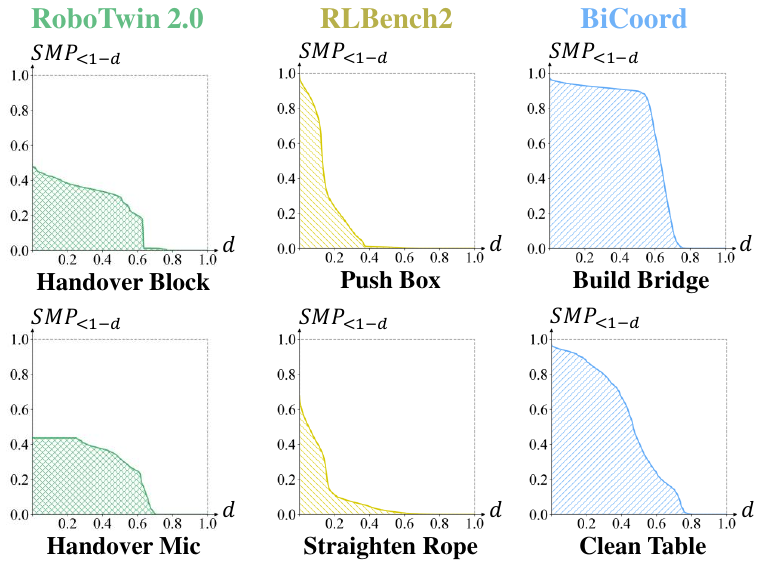}
  }
  \vspace{-4mm}
  \caption{\textbf{The STI curves.} Tasks in BiCoord achieve substantially higher STI values than those in existing benchmarks.}
  \vspace{-3mm}
  \label{fig:sti}
\end{figure}
We conduct a quantitative analysis with existing bimanual manipulation benchmarks, \eg, RoboTwin 2.0~\cite{chen2025robotwin} and RLBench2~\cite{grotz2024peract2}.

As shown in \cref{tab:comparison}, SMP of RLBench2 reaches $97.10\%$, which means the dual arms perform actions simultaneously at nearly all timesteps. However, with an ARD of $114.93\%$, RLBench2 is weak in space cooperation, since dual arms are far from each other. In comparison, RoboTwin 2.0 features a lower ARD and an equivalent MRD, indicating stronger spatial correlation. However, SMP is only $26.10\%$ in RoboTwin 2.0, showing a lack of temporal synergy. Due to deficiency in either space or time, both benchmarks demonstrate restricted STI. Such a feature can also be observed from \cref{fig:sti}, where STI curve of RoboTwin 2.0 is closer to space axis ($d$), while that of RLBench2 is closer to time axis ($SMP_{<1-d}$).

For long-horizon metrics, as shown in Tab.~\ref{tab:comparison}, the tasks in both benchmarks have an average of $1$-$2$ stages and $2$–$3$ objects, with an average trajectory length of around $200$ timesteps. Such statistics show that they mainly focus on short-horizon manipulation tasks.

\begin{figure}[t]
  \centering 
  \resizebox{1\linewidth}{!}{
  \includegraphics{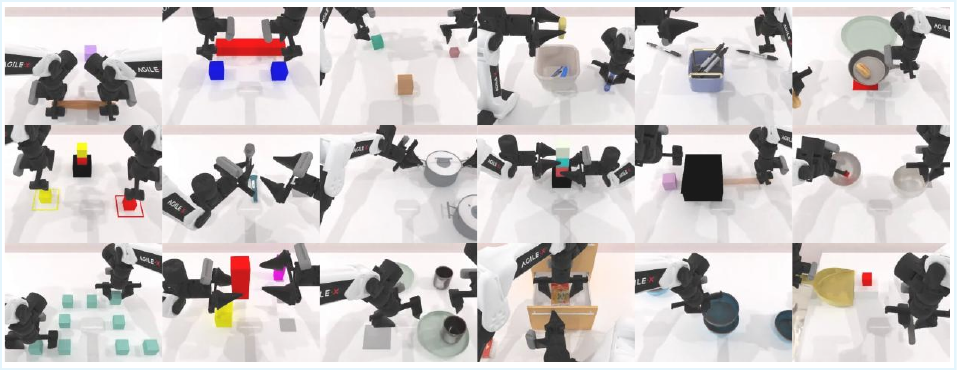}
  }
  \vspace{-5mm}
  \caption{\textbf{Tasks in BiCoord.} Each task requires high-level spatial-temporal coordination within a long manipulation.}
  \vspace{-2mm}
  \label{fig:tasks}
\end{figure}

\section{BiCoord}
\label{sec:data}
\begin{figure}[!htpb]
  \centering 
  \resizebox{1\linewidth}{!}{
  \includegraphics{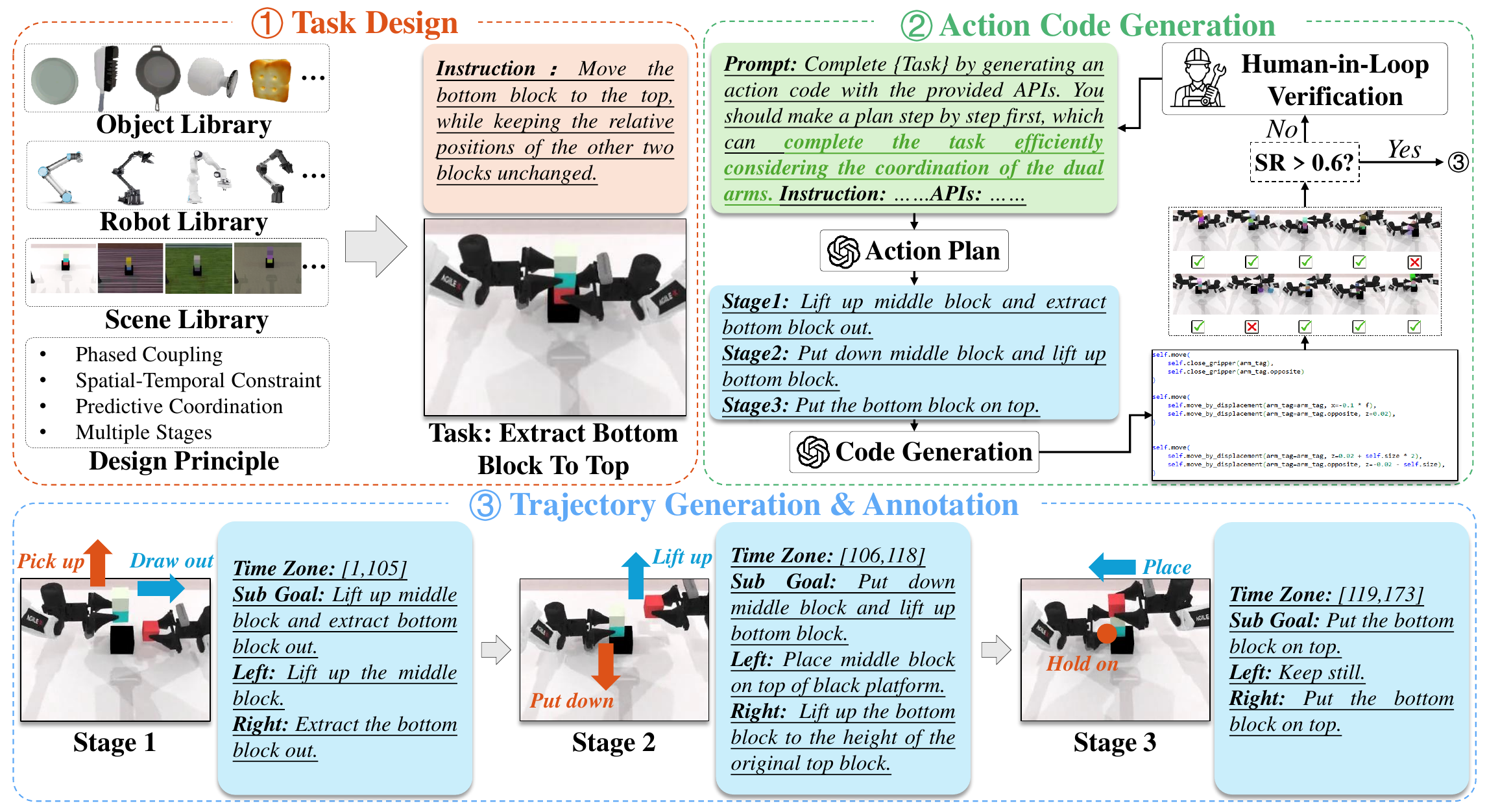}
  }
  \vspace{-5mm}
  \caption{\textbf{Pipeline for building BiCoord.}}
  \vspace{-5mm}
  \label{fig:pipeline}
\end{figure}

As shown in \cref{sec:met:eval}, existing bimanual manipulation benchmarks are deficient in coordination and length. To address this, we propose BiCoord, a bimanual manipulation benchmark requiring both high-level coordination and long-term inference, as shown in \cref{fig:tasks}. In the following, we will present the building pipeline of BiCoord in \cref{sec:data:pipe}, and introduce its features in \cref{sec:data:feat}.

\subsection{Pipeline for Building BiCoord}
\label{sec:data:pipe}

We build BiCoord on the basis of RoboTwin 2.0~\cite{chen2025robotwin}. As shown in \cref{fig:pipeline}, the building of BiCoord includes three stages.

\noindent\textbf{Task Design.} We design tasks based on objects in the RoboTwin-OD~\cite{mu2025robotwin,chen2025robotwin}. All tasks are carefully designed with high-level bimanual coordination and multiple manipulation stages. Besides, multi-type embodiments and diverse scene textures are also supported.  

\noindent\textbf{Action Code Generation.} We first prompt the coding agent with tasks details and API list to generate action code. The code is then tested in simulation on $10$ random seeds, whose success rate needs to surpass $0.6$. Otherwise, human-in-loop verification is introduced to confirm the weakness and guide the next-step generation.

\noindent\textbf{Trajectory Generation \& Annotation.} Qualified action codes are used for trajectory generation, with 100 successful trajectories for each task. Meanwhile, stage-wise annotations can be automatically obtained, including time zones, sub-goals and arm behaviours.

\subsection{Features of BiCoord}
\label{sec:data:feat}
BiCoord is focused on spatial-temporal coordination and long-horizon manipulation, with stage-wise annotation and evaluation for fine-grained policy training and testing.

\noindent\textbf{Tight Spatio-Temporal Coordination.} Tasks in BiCoord impose high requirements on the dual-arm coordination in both space and time. As shown in \cref{tab:comparison}, the SMP of BiCoord reaches $92.81\%$, indicating that the dual arms act simultaneously in nearly all timesteps. Besides, MRD and ARD are decreased by $45.78\%$ and $32.29\%$ compared to previous benchmarks, showing that dual arms can collaborate within a closer spatial range in BiCoord. Moreover, time and space are highly coupled in BiCoord, with the STI up to $42.16\%$, which is nearly $4$x times of previous benchmarks. Such a couple could also be observed from \cref{fig:sti}.

\noindent\textbf{Long-Horizon Manipulation.} The task length of BiCoord is longer than that of previous benchmarks. A task in BiCoord contains an average of $4.27$ stages, which is nearly $3$x times compared to previous benchmarks. Besides, the average trajectory length and object number are improved by $63.35\%$ and $64.13\%$ respectively. 

\noindent\textbf{Stage-Wise Annotation and Evaluation.} To support fine-grained policy training and testing, BiCoord provides stage-wise annotation and evaluation. For training, each trajectory is divided into several stages, where each stage is coupled with a sub-goal and arm behaviours. For testing, BiCoord verifies the achievement of sub-goals besides final result, providing more granular details about the manipulation process. Moreover, each stage is assigned a score $s\in [0,1]$, with the sum of all stage scores equal to $1$. Based on this, we propose \textbf{Stage-Wise Success Rate (SSR)}:
\begin{equation}
    SSR = \frac{1}{N}\sum_{i=1}^N\sum_{j=1}^Ms_{ij}\cdot c_{ij},
\end{equation}
where $N$ is the number of test rollouts, $M$ is the number of task stages, $s_{ij}$ is assigned stage score, $c_{ij}\in\{0,1\}$ indicates whether the stage is successfully completed.

\begin{figure}[t]
  \centering 
  \resizebox{0.45\textwidth}{!}{
  \includegraphics{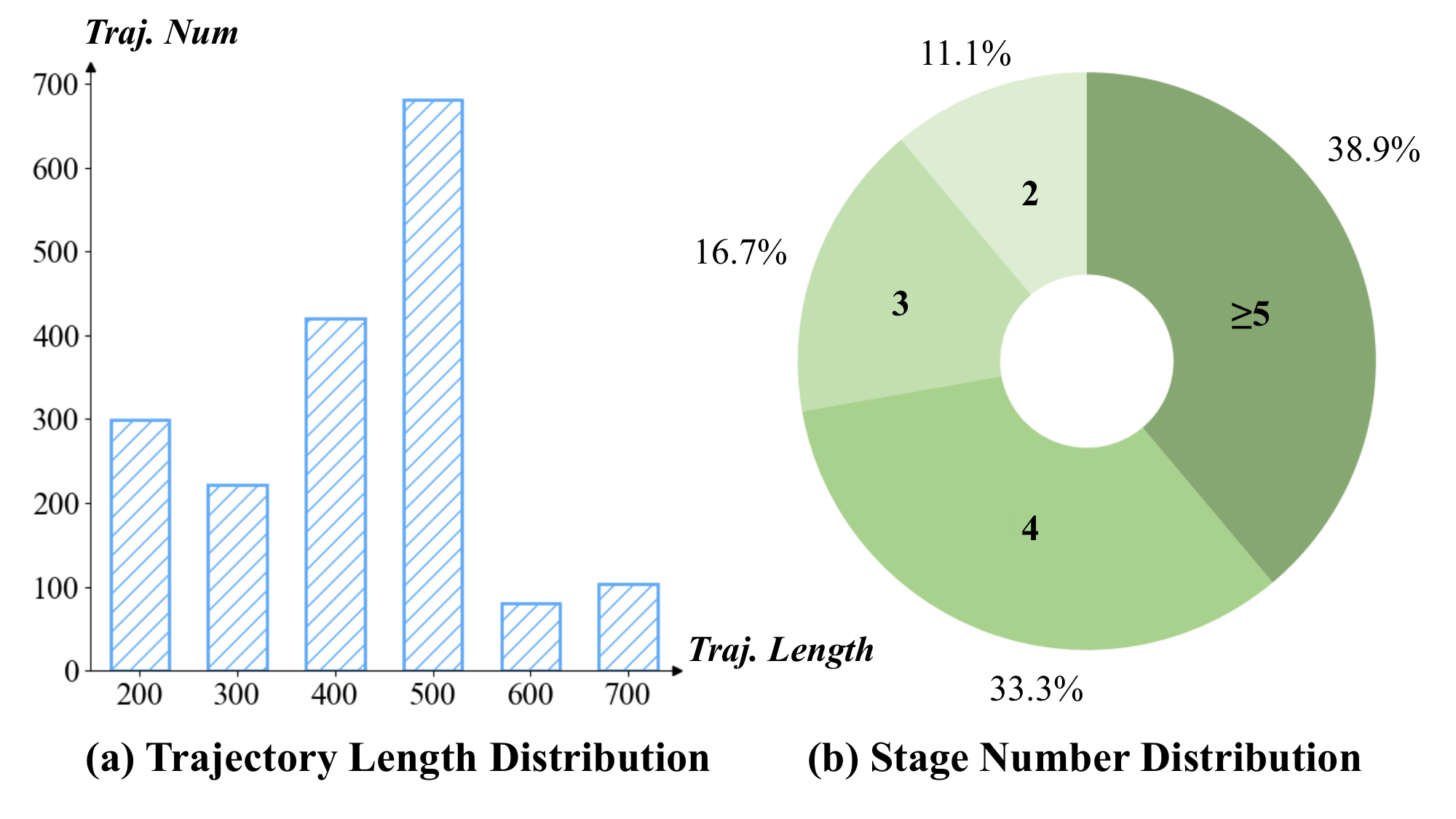}
  }
  \caption{\textbf{Statistics of BiCoord.}}
  \label{fig:statistics}
\end{figure}
\section{Experiments}
\label{sec:exp}

\subsection{Experiment Setups}
We evaluate Diffusion Policy (DP)~\cite{chi2023diffusion} and three VLAs (RDT~\cite{liu2024rdt1b}, OpenVLA-OFT~\cite{openvla_oft}, Pi0~\cite{black2024pi_0}) on BiCoord. For single-task models, all policies are trained referring configs in RoboTwin 2.0~\cite{chen2025robotwin}, which are detailed in Supplements. For multi-task models, all policies are finetuned for $50,000$ steps from official weights with a total batch size of $32$. All experiments are conducted on A800 40GB GPUs.

For evaluation, all policies are evaluated on $100$ rollouts for each task. Besides success rate (SR), we also report stage-wise success rate (SSR) and trajectory length (TL). Note that TL is computed on successful trajectories. For fair comparison, when calculating the average TL ov all tasks, the tasks with any zero SR will be excluded.  

\subsection{Results of Single-Task Learning}

\begin{table}[t]
\centering
\caption{\textbf{Results of single-task learning.} VLA models exhibit higher SR/SSR, which DP demonstrates a lower TL.}
\vspace{-2mm}
\resizebox{1\linewidth}{!}{
{
\renewcommand{\arraystretch}{1.35}
\setlength{\tabcolsep}{5pt}
\begin{tabular}{r||ccc||ccc||ccc||ccc}
\hline

 \rowcolor[HTML]{f8f9fa} \multicolumn{1}{c||}{}& \multicolumn{3}{c||}{DP} & \multicolumn{3}{c||}{RDT} & \multicolumn{3}{c||}{OpenVLA-OFT} & \multicolumn{3}{c}{Pi0} \\
\rowcolor[HTML]{f8f9fa} \multicolumn{1}{c||}{\multirow{-2}{*}{Task}} & SR & SSR & TL & SR & SSR & TL & SR & SSR & TL & SR & SSR & TL \\
\hline
\hline
Balance Roller & 34.0 & 67.0 & \textbf{101} & 39.0 & 69.5 & 128 & \textbf{89.0} & \textbf{94.5} & 124 & 62.0 & 79.5 & 149 \\
Build Bridge & 68.0 & 69.5 & \textbf{243} & \textbf{76.0} & \textbf{78.5} & 275 & 66.0 & 67.8 & 282 & 64.0 & 65.0 & 291 \\
Build Tower With Blocks & 0.0 & 1.5 & - & 0.0 & 0.5 & - & 0.0 & \textbf{7.0} & - & 0.0 & 1.5 & - \\
Clean Table & 0.0 & 28.5 & - & 0.0 & 23.8 & - & 1.0 & 33.0 & 775 & \textbf{9.0} & \textbf{46.8} & \textbf{672} \\
Collect Pens & 1.0 & 46.8 & \textbf{378} & 57.0 & \textbf{87.8} & 487 & 35.0 & 75.8 & 456 & \textbf{68.0} & 85.8 & 414 \\
Cook & 8.0 & 31.2 & \textbf{398} & 27.0 & 48.2 & 506 & 26.0 & 33.5 & 457 & \textbf{35.0} & \textbf{57.2} & 549 \\
Divide Block Tower & 7.0 & \textbf{28.9} & \textbf{342} & 2.0 & 23.2 & 384 & 3.0 & 14.6 & 367 & \textbf{9.0} & 28.0 & 378 \\
Exchange Mics & 58.0 & 77.0 & \textbf{371} & \textbf{74.0} & \textbf{78.5} & 458 & 68.0 & 68.0 & 425 & 52.0 & 68.0 & 444 \\
Exchange Pots & 90.0 & 93.5 & \textbf{386} & 81.0 & 87.0 & 448 & \textbf{100.0} & \textbf{100.0} & 450 & 96.0 & 97.0 & 450 \\
Extract Bottom Block To Top & 80.0 & 80.5 & \textbf{151} & 72.0 & 74.5 & 192 & 75.0 & 75.5 & 175 & \textbf{83.0} & \textbf{84.5} & 200 \\
Fetch Block With Roller & 4.0 & 52.0 & 228 & 92.0 & 96.0 & 256 & \textbf{94.0} & \textbf{96.5} & \textbf{225} & 93.0 & 96.0 & 250 \\
Handover Block With Bowls & 1.0 & 49.5 & \textbf{150} & 0.0 & 45.0 & - & 0.0 & 19.0 & - & \textbf{2.0} & \textbf{50.5} & \textbf{150} \\
Jigsaw & 0.0 & 0.8 & - & \textbf{44.0} & \textbf{79.8} & \textbf{406} & 4.0 & 21.0 & 525 & 17.0 & 49.8 & 412 \\
Match Blocks With Signs & 2.0 & 16.0 & \textbf{561} & 12.0 & 30.3 & 640 & 5.0 & 7.3 & 715 & \textbf{20.0} & \textbf{34.3} & 615 \\
Place Plate And Cup & \textbf{71.0} & \textbf{89.2} & \textbf{277} & 51.0 & 78.2 & 320 & 61.0 & 80.5 & 306 & 56.0 & 82.0 & 304 \\
Put Objects Cabinet & 67.0 & \textbf{81.0} & \textbf{440} & 56.0 & 70.5 & 529 & \textbf{70.0} & 80.5 & 477 & 65.0 & 75.5 & 501 \\
Stack Bowls & \textbf{21.0} & 41.3 & \textbf{388} & 15.0 & \textbf{48.0} & 533 & 10.0 & 44.3 & 422 & 14.0 & 35.0 & 439 \\
Sweep Block & 83.0 & 83.0 & \textbf{198} & 13.0 & 13.0 & 256 & 22.0 & 22.0 & 242 & \textbf{91.0} & \textbf{91.0} & 249 \\

\hline
Average & 33.1 & 52.1 & \textbf{307} & 39.5 & 57.3 & 387 & 40.5 & 52.3 & 401 & \textbf{46.4} & \textbf{62.6} & 380 \\
\hline
\end{tabular}
}
}
\label{tab:single_task}
\end{table}

The results of single-task learning are presented in \cref{tab:single_task}. 

\noindent\textbf{Outstanding performance of VLA models.} With pretrained knowledge, VLA models achieve better performance compared to DP. Though RDT, Pi0 and OpenVLA-OFT adopt different model architectures, they all benefit from large-scale pretraining, showing better ability in handling complex tasks. Such a phenomenon indicates that large-scale pre-training is also meaningful in the field of embodied intelligence, just like in vision-language models.

\noindent\textbf{High efficiency of DP.} DP generally takes fewer timesteps to complete a task compared to VLA models. For example, on \textit{Place Plate And Cup}, DP achieves the highest SR of $71\%$, while it only takes an average of $277$ steps, which is $9.75\%$ shorter than that of Pi0. This inspires a potential to combine the efficiency of DP with the priors of VLA.

\begin{figure}[t]
  \centering 
  \resizebox{0.48\textwidth}{!}{
  \includegraphics{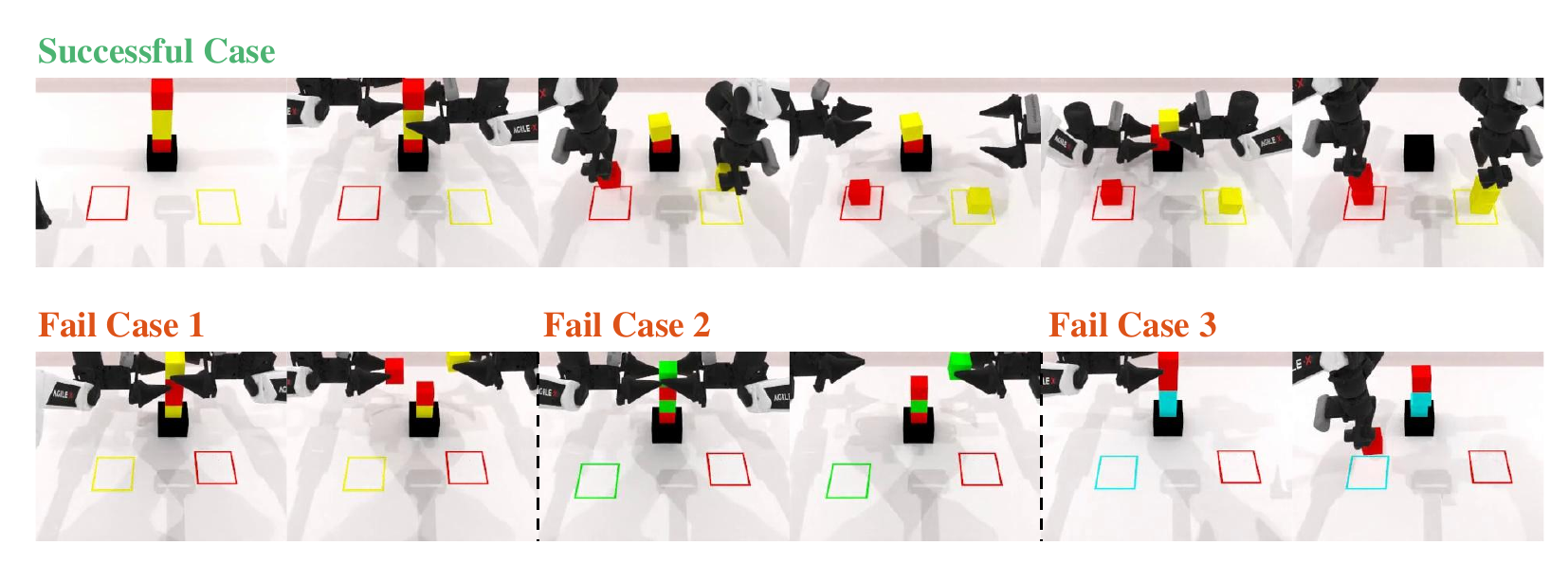}
  }
  \caption{\textbf{Visualizations of Pi0 on \textit{Divide Block Tower} task.} Grasping errors occur when the color and order of the blocks change, demonstrating limited reasoning ability.}
  \label{fig:reason}
\end{figure}

\noindent\textbf{Poor Reasoning Ability.} Current policies are relatively lacking in reasoning capability and struggle to adapt to changes in initial conditions. For example, Pi0 possesses the ability to accomplish \textit{Divide Block Tower} task, as the successful case in \cref{fig:reason} shows. However, when colors and order of the blocks change, the model fails to grasp the blocks to the correct side. Since the training trajectories already include various block colors and sequences, this phenomenon indicates the model cannot associate the blocks with landmarks through colors, showing a lack in reasoning capability. This dilemma may arise from rigid imitation learning. Reinforcement learning and high-level planning are likely to alleviate this issue.

\begin{figure}[t]
  \centering 
  \resizebox{0.48\textwidth}{!}{
  \includegraphics{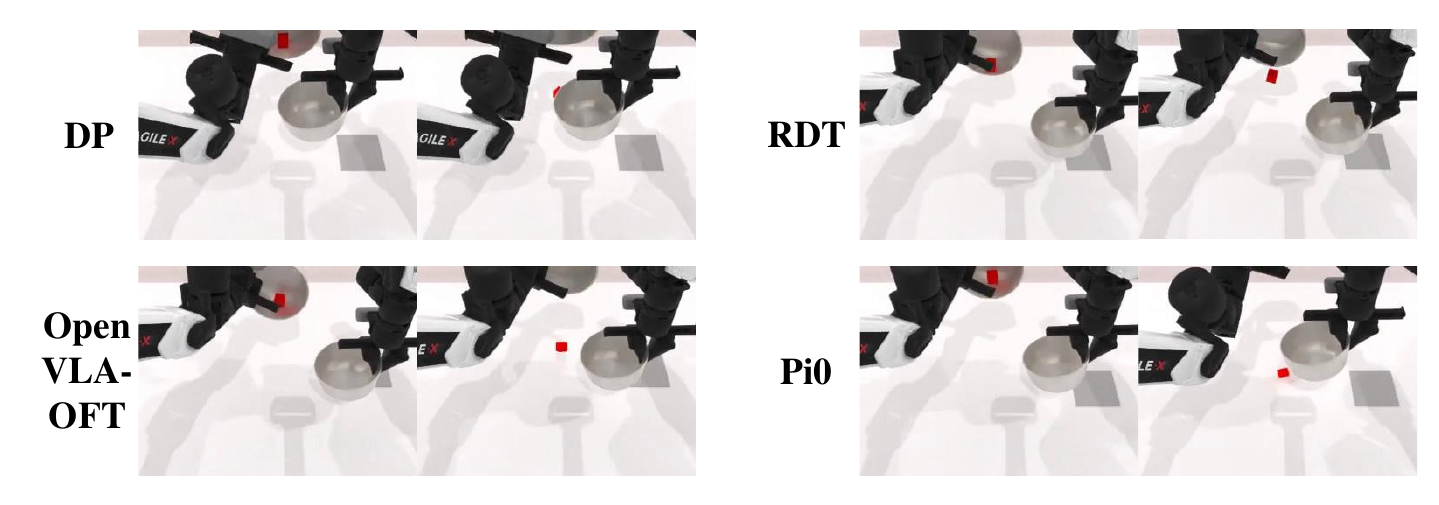}
  }
  \caption{\textbf{Visualizations on \textit{Handover Block With Bowls} task.} The block is poured out before the two bowls completely aligned, showing weak abilities in precise alignment.}
  \label{fig:precise}
\end{figure}

\noindent\textbf{Deficiency in precise alignment.} Current policies exhibit deficiency in precise manipulation. For example, \textit{Handover Block With Bowls} requires pouring the block from one bowl into the other, demanding precise alignment. As shown in \cref{tab:single_task}, all policies achieve nearly zero SR on this task, demonstrating poor alignment ability. Failing reason is shown in ~\cref{fig:precise}, where pouring starts before the bowls are aligned. Interestingly, DP seems to be more accurate than VLA models when aligning two bowls. Researching how to improve VLAs' ability for precise alignment is worthwhile in the future.

\begin{figure}[t]
  \centering 
  \resizebox{1\linewidth}{!}{
  \includegraphics{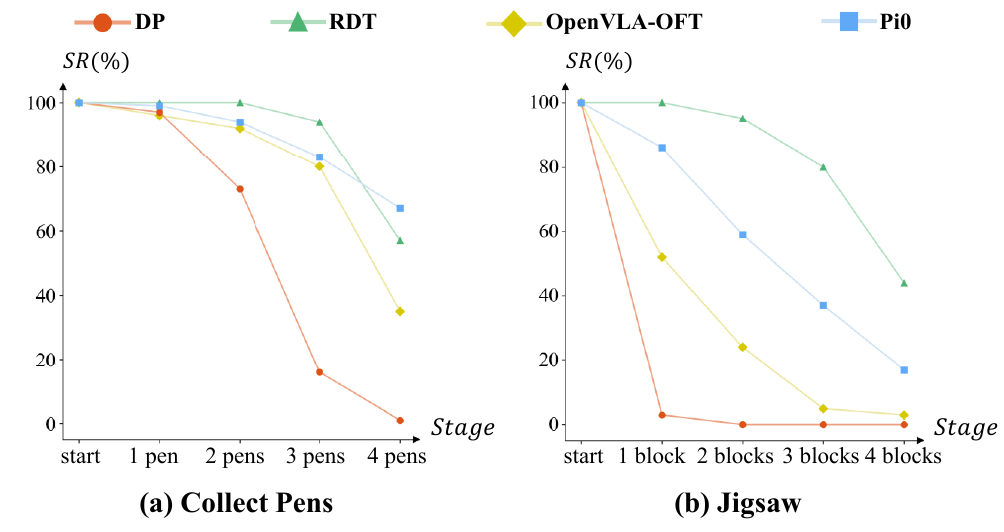}
  }
  \vspace{-5mm}
  \caption{\textbf{Stage-wise analysis.} We present two examples here to demonstrate stage-wise evaluation.}
  \vspace{-2mm}
  \label{fig:stage-wise}
\end{figure}

\noindent\textbf{Stage-wise analysis.} Thanks to stage-wise evaluation, we can conduct a detailed analysis of the policy’s performance at each stage. For example, in \cref{fig:stage-wise}(a), dual arms need to place four pens into the pen holder on \textit{Collect Pens} task. Though RDT and OpenVLA-OFT performs well at early stages, the SR greatly drops when placing the fourth pen. In contrast, Pi0 generally maintains a high SR through the whole process, showing a better long-horizon ability in this task. Similarly, as shown in \cref{fig:stage-wise}(b), with the number of blocks increasing, the SR of all policies drops sharply. Such a result indicates that current policies are still weak in long-horizon manipulation.

\begin{figure}[!htpb]
  \centering 
  \includegraphics[width=1\linewidth]{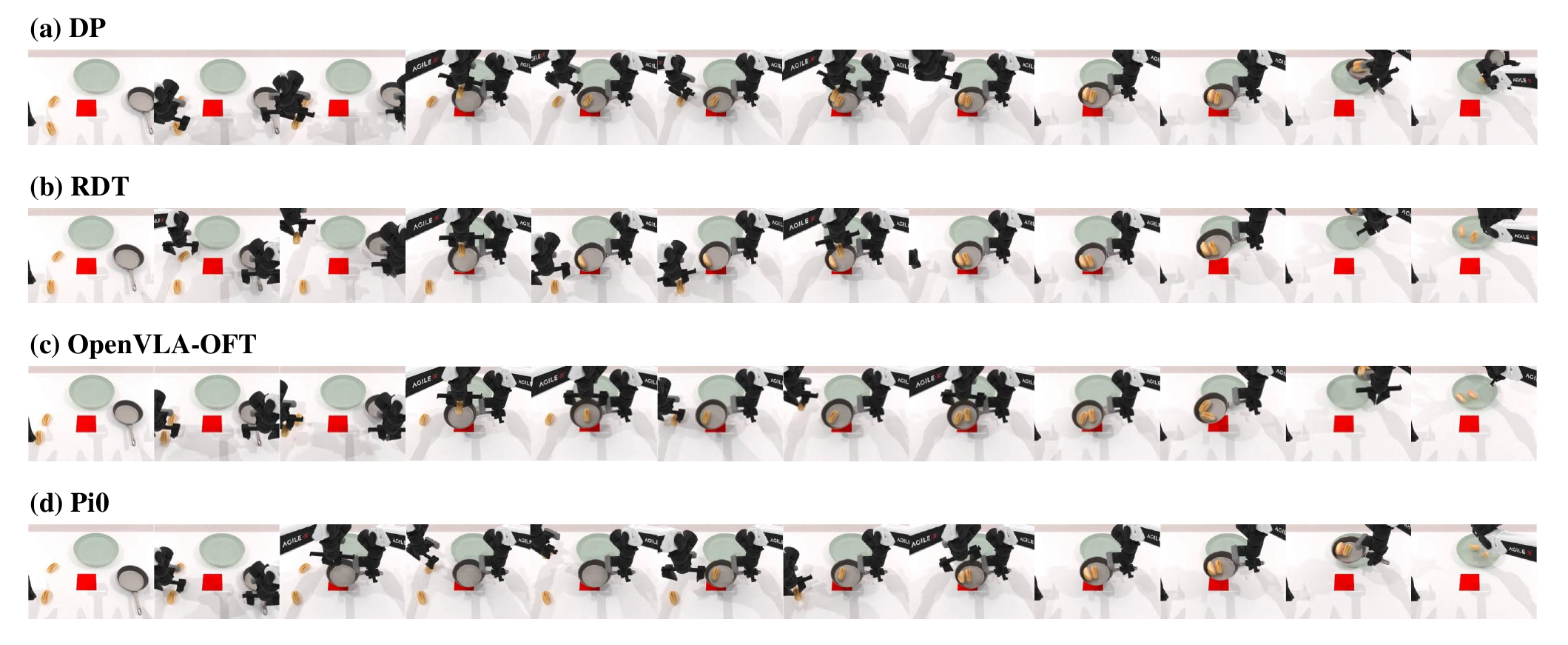}
  \vspace{-7mm}
  \caption{\textbf{Visualizations on \textit{Cook} task.} 
  }
    \vspace{-2mm}
  \label{fig:viz}
\end{figure}

\subsection{Results of Multi-Task Learning}

\begin{table}[t]
\centering
\caption{\textbf{Results of multi-task learning.} The SR of multi-task models greatly drops compared to single-task ones.}
\vspace{-2mm}
\resizebox{1\linewidth}{!}{
{
\renewcommand{\arraystretch}{1.35}
\setlength{\tabcolsep}{5pt}
\begin{tabular}{r||ccc||ccc||ccc}
\hline

 \rowcolor[HTML]{f8f9fa} \multicolumn{1}{c||}{}& \multicolumn{3}{c||}{RDT} & \multicolumn{3}{c||}{OpenVLA-OFT} & \multicolumn{3}{c}{Pi0} \\
\rowcolor[HTML]{f8f9fa} \multicolumn{1}{c||}{\multirow{-2}{*}{Task}} & SR & SSR & TL & SR & SSR & TL & SR & SSR & TL \\
\hline
\hline
Balance Roller & 15.0 & 43.0 & 149 & 49.0 & 74.5 & \textbf{137} & \textbf{68.0} & \textbf{81.5} & 151 \\
Build Bridge & \textbf{47.0} & \textbf{50.2} & 370 & 2.0 & 2.0 & 312 & 44.0 & 47.0 & \textbf{289} \\
Build Tower With Blocks & 0.0 & 1.0 & - & 0.0 & 0.0 & - & \textbf{2.0} & \textbf{5.0} & \textbf{650} \\
Clean Table & 0.0 & 17.5 & - & 2.0 & 25.0 & \textbf{612} & \textbf{6.0} & \textbf{46.2} & 658 \\
Collect Pens & 15.0 & 68.8 & 495 & 5.0 & 46.2 & 795 & \textbf{47.0} & \textbf{81.8} & \textbf{457} \\
Cook & 9.0 & \textbf{31.0} & 569 & 10.0 & 25.0 & \textbf{442} & \textbf{14.0} & 29.5 & 504 \\
Divide Block Tower & \textbf{2.0} & \textbf{12.8} & \textbf{384} & 0.0 & 11.1 & - & 0.0 & 10.8 & - \\
Exchange Mics & 23.0 & 35.0 & 467 & \textbf{66.0} & \textbf{67.5} & \textbf{418} & 52.0 & 59.5 & 446 \\
Exchange Pots & \textbf{92.0} & \textbf{96.0} & 459 & 53.0 & 56.0 & \textbf{442} & 52.0 & 60.5 & 476 \\
Extract Bottom Block To Top & 17.0 & 25.5 & 192 & \textbf{53.0} & \textbf{55.5} & \textbf{177} & 40.0 & 43.5 & 208 \\
Fetch Block With Roller & 0.0 & 49.5 & - & \textbf{44.0} & \textbf{70.5} & \textbf{225} & 43.0 & 69.5 & 250 \\
Handover Block With Bowls & 0.0 & \textbf{20.0} & - & 0.0 & 3.0 & - & \textbf{4.0} & 9.0 & \textbf{188} \\
Jigsaw & 0.0 & 11.5 & - & 0.0 & 25.2 & - & \textbf{6.0} & \textbf{36.5} & \textbf{400} \\
Match Blocks With Signs & 1.0 & 7.0 & 1216 & 5.0 & 8.7 & 760 & \textbf{8.0} & \textbf{18.7} & \textbf{638} \\
Place Plate And Cup & 40.0 & 70.0 & 320 & \textbf{55.0} & \textbf{79.2} & 315 & 25.0 & 66.2 & \textbf{302} \\
Put Objects Cabinet & \textbf{31.0} & 49.0 & 526 & 26.0 & \textbf{49.5} & \textbf{486} & 6.0 & 24.0 & 575 \\
Stack Bowls & 3.0 & \textbf{28.3} & 597 & 1.0 & 14.0 & \textbf{400} & \textbf{12.0} & 28.0 & 471 \\
Sweep Block & 10.0 & 10.0 & 256 & 44.0 & 44.0 & \textbf{234} & \textbf{61.0} & \textbf{61.0} & 253 \\

\hline
Average & 16.9 & 34.8 & 461 & 23.1 & 36.5 & 411 & \textbf{27.2} & \textbf{43.2} & \textbf{406} \\
\hline
\end{tabular}
}
}
\label{tab:multi_task}
\end{table}

Besides single-task setting, we also conduct tests in multi-task setting, where success rate generally drops compared to single-task models. For example, the SR of Pi0 drops from $46.4\%$ to $27.2\%$. Such results are consistent with expectations, since completing multiple tasks with a single model is more challenging. However, it is a surprise to see that multi-task models also achieve improvements on certain tasks. As shown in \cref{tab:single_task} and \cref{tab:multi_task}, OpenVLA-OFT's SR on \textit{Sweep Block} is improved from $22.0\%$ to $44.0\%$. In fact, although different tasks may compete with each other within the same model, their shared skills may also promote each other. By designing methods to explore the commonalities across different tasks, VLAs may be endowed with stronger generalization capabilities.

\subsection{Full-Trajectory Visualization}

To intuitively compare different policies, we present the full trajectory of each policy on the \textit{Cook} task in \cref{fig:viz}.

\noindent\textbf{Recovery ability.} As shown in \cref{fig:viz}(d), Pi0 fails to grasp the first bread at beginning. However, instead of proceeding with the subsequent operations, Pi0 tries to pick up the bread again after placing the other bread. This indicates that VLAs like Pi0 do not rigidly adhere to trained trajectories. Instead, they potentially possess the ability to evaluate the task state and correct previous errors.

\noindent\textbf{Stability.} As shown in \cref{fig:viz}(b) and \cref{fig:viz}(c), RDT and OpenVLA-OFT do not hold the skillet stably, where the skillet is tilted. In contrast, DP can place the skillet horizontally. Such a phenomenon indicates that DP’s manipulation is more stable. In fact, by analyzing more cases, we find that DP’s failures are mainly caused by its inability to flexibly adjust the grasping pose when picking breads of various poses. In contrast, VLAs' failures are mainly caused by unstable skillet grasping, where breads may fall out from the skillet. How to combine the stability of DP with the generalization of VLA is a problem worthy of future research.

\noindent\textbf{Efficiency.} When pouring breads into the skillet, DP tends to move closer to the plate and adopt smaller-range movements. In contrast, VLAs tend to perform rotations with larger amplitudes. Such a feature also shows in TL metric, where DP takes $12.91\%$ fewer timesteps to complete $Cook$ task (as shown in \cref{tab:single_task}).
\section{Conclusion}
In this paper, we propose BiCoord, a bimanual manipulation benchmark towards long-horizon spatial-temporal coordination. We propose spatial-temporal metrics, laying a theoretical foundation for bimanual manipulation tasks. Based on this, we release BiCoord, which is designed to facilitate research on long-horizon spatial-temporal coordination. To support fine-grained policy training and testing, BiCoord provides stage-wise annotation and evaluation. We baseline four popular manipulation policies on BiCoord, and discuss on interesting phenomena observed from the testing results. We hope BiCoord could facilitate related research in the future.
\label{sec:con}

\bibliographystyle{ACM-Reference-Format}
\bibliography{sample-base}

@String(CVPR= {IEEE Conf. Comput. Vis. Pattern Recog.})

@String(CVPR  = {CVPR})

@inproceedings{adaptdiffuser,
  title={AdaptDiffuser: Diffusion Models as Adaptive Self-evolving Planners},
  author={Liang, Zhixuan and Mu, Yao and Ding, Mingyu and Ni, Fei and Tomizuka, Masayoshi and Luo, Ping},
  booktitle={International Conference on Machine Learning},
  pages={20725--20745},
  year={2023},
  organization={PMLR}
}

@String(CVPR  = {IEEE Conf. Comput. Vis. Pattern Recog.})

@String(NeurIPS = {Adv. Neural Inform. Process. Syst.})

@String(NeurIPS = {NeurIPS})

@article{ze20243d,
  title={3d diffusion policy},
  author={Ze, Yanjie and Zhang, Gu and Zhang, Kangning and Hu, Chenyuan and Wang, Muhan and Xu, Huazhe},
  journal={arXiv preprint arXiv:2403.03954},
  year={2024}
}

@inproceedings{walke2023bridgedata,
  title={Bridgedata v2: A dataset for robot learning at scale},
  author={Walke, Homer Rich and Black, Kevin and Zhao, Tony Z and Vuong, Quan and Zheng, Chongyi and Hansen-Estruch, Philippe and He, Andre Wang and Myers, Vivek and Kim, Moo Jin and Du, Max and others},
  booktitle={Conference on Robot Learning},
  pages={1723--1736},
  year={2023},
  organization={PMLR}
}

@article{han2025robocerebra,
  title={RoboCerebra: A Large-scale Benchmark for Long-horizon Robotic Manipulation Evaluation},
  author={Han, Songhao and Qiu, Boxiang and Liao, Yue and Huang, Siyuan and Gao, Chen and Yan, Shuicheng and Liu, Si},
  journal={arXiv preprint arXiv:2506.06677},
  year={2025}
}

@article{fu2024mobile,
  title={Mobile aloha: Learning bimanual mobile manipulation with low-cost whole-body teleoperation},
  author={Fu, Zipeng and Zhao, Tony Z and Finn, Chelsea},
  journal={arXiv preprint arXiv:2401.02117},
  year={2024}
}

@article{zhao2023learning,
  title={Learning fine-grained bimanual manipulation with low-cost hardware},
  author={Zhao, Tony Z and Kumar, Vikash and Levine, Sergey and Finn, Chelsea},
  journal={arXiv preprint arXiv:2304.13705},
  year={2023}
}

@article{mees2022calvin,
  title={Calvin: A benchmark for language-conditioned policy learning for long-horizon robot manipulation tasks},
  author={Mees, Oier and Hermann, Lukas and Rosete-Beas, Erick and Burgard, Wolfram},
  journal={IEEE Robotics and Automation Letters},
  volume={7},
  number={3},
  pages={7327--7334},
  year={2022},
  publisher={IEEE}
}

@inproceedings{khazatsky2024droid,
  title={DROID: A large-scale in-the-wild robot manipulation dataset},
  author={Khazatsky, Alexander and Pertsch, Karl and Nair, Suraj and Balakrishna, Ashwin and Dasari, Sudeep and Karamcheti, Siddharth and Nasiriany, Soroush and Srirama, Mohan Kumar and Chen, Lawrence Yunliang and Ellis, Kirsty and others},
  booktitle={Robotics: Science and Systems},
  year={2024}
}

@article{ebert2021bridge,
  title={Bridge data: Boosting generalization of robotic skills with cross-domain datasets},
  author={Ebert, Frederik and Yang, Yanlai and Schmeckpeper, Karl and Bucher, Bernadette and Georgakis, Georgios and Daniilidis, Kostas and Finn, Chelsea and Levine, Sergey},
  journal={arXiv preprint arXiv:2109.13396},
  year={2021}
}

@article{mittal2023orbit,
  title={Orbit: A unified simulation framework for interactive robot learning environments},
  author={Mittal, Mayank and Yu, Calvin and Yu, Qinxi and Liu, Jingzhou and Rudin, Nikita and Hoeller, David and Yuan, Jia Lin and Singh, Ritvik and Guo, Yunrong and Mazhar, Hammad and others},
  journal={IEEE Robotics and Automation Letters},
  volume={8},
  number={6},
  pages={3740--3747},
  year={2023},
  publisher={IEEE}
}

@article{james2020rlbench,
  title={Rlbench: The robot learning benchmark \& learning environment},
  author={James, Stephen and Ma, Zicong and Arrojo, David Rovick and Davison, Andrew J},
  journal={IEEE Robotics and Automation Letters},
  volume={5},
  number={2},
  pages={3019--3026},
  year={2020},
  publisher={IEEE}
}

@inproceedings{todorov2012mujoco,
  title={Mujoco: A physics engine for model-based control},
  author={Todorov, Emanuel and Erez, Tom and Tassa, Yuval},
  booktitle={IEEE/RSJ International Conference on Intelligent Robots and Systems},
  pages={5026--5033},
  year={2012},
}

@inproceedings{robosuite2020,
  title={robosuite: A Modular Simulation Framework and Benchmark for Robot Learning},
  author={Yuke Zhu and Josiah Wong and Ajay Mandlekar and Roberto Mart{\'i}n-Mart{\'i}n},
  booktitle={arXiv preprint arXiv:2009.12293},
  year={2020}
}

@article{dasari2021rb2,
  title={RB2: Robotic Manipulation Benchmarking with a Twist},
  author={Dasari, Sudeep and Wang, Jianren and Hong, Joyce and Bahl, Shikhar and Lin, Yixin and Wang, Austin S and Thankaraj, Abitha and Singh Chahal, Karanbir and Calli, Berk and Gupta, Saurabh and others},
  journal={NeurIPS 2021 Datasets and Benchmarks Track},
  year={2021}
}

@inproceedings{wang2022bulletarm,
  title={Bulletarm: An open-source robotic manipulation benchmark and learning framework},
  author={Wang, Dian and Kohler, Colin and Zhu, Xupeng and Jia, Mingxi and Platt, Robert},
  booktitle={The International Symposium of Robotics Research},
  pages={335--350},
  year={2022},
  organization={Springer}
}

@article{chatzilygeroudis2020benchmark,
  title={Benchmark for bimanual robotic manipulation of semi-deformable objects},
  author={Chatzilygeroudis, Konstantinos and Fichera, Bernardo and Lauzana, Ilaria and Bu, Fanjun and Yao, Kunpeng and Khadivar, Farshad and Billard, Aude},
  journal={IEEE Robotics and Automation Letters},
  volume={5},
  number={2},
  pages={2443--2450},
  year={2020},
  publisher={IEEE}
}

@inproceedings{fan2023arctic,
  title={ARCTIC: A dataset for dexterous bimanual hand-object manipulation},
  author={Fan, Zicong and Taheri, Omid and Tzionas, Dimitrios and Kocabas, Muhammed and Kaufmann, Manuel and Black, Michael J and Hilliges, Otmar},
  booktitle={Proceedings of the IEEE/CVF conference on computer vision and pattern recognition},
  pages={12943--12954},
  year={2023}
}

@article{aldaco2024aloha,
  title={ALOHA 2: An Enhanced Low-Cost Hardware for Bimanual Teleoperation},
  author={Aldaco, Jorge and Armstrong, Travis and Baruch, Robert and Bingham, Jeff and Chan, Sanky and Draper, Kenneth and Dwibedi, Debidatta and Finn, Chelsea and Florence, Pete and Goodrich, Spencer and others},
  journal={arXiv preprint arXiv:2405.02292},
  year={2024}
}

@article{sferrazza2024humanoidbench,
  title={Humanoidbench: Simulated humanoid benchmark for whole-body locomotion and manipulation},
  author={Sferrazza, Carmelo and Huang, Dun-Ming and Lin, Xingyu and Lee, Youngwoon and Abbeel, Pieter},
  journal={arXiv preprint arXiv:2403.10506},
  year={2024}
}

@article{taomaniskill3,
  title={ManiSkill3: GPU Parallelized Robotics Simulation and Rendering for Generalizable Embodied AI},
  author={Stone Tao and Fanbo Xiang and Arth Shukla and Yuzhe Qin and Xander Hinrichsen and Xiaodi Yuan and Chen Bao and Xinsong Lin and Yulin Liu and Tse-kai Chan and Yuan Gao and Xuanlin Li and Tongzhou Mu and Nan Xiao and Arnav Gurha and Zhiao Huang and Roberto Calandra and Rui Chen and Shan Luo and Hao Su},
  journal={arXiv preprint arXiv:2410.00425},
  year={2024}
}

@article{xiang2020sapien,
  title={SAPIEN: A SimulAted Part-based Interactive ENvironment},
  author={Fanbo Xiang and He Wang and Yuzhe Qin and Austin Wang and Hejia Zhang and Yikuan Xia and Binbin Lin and Yuzhe Wu and Chengcheng Tang and Yixin Zhu and Li Yi and Leonidas J. Guibas and Hao Su},
  journal={Proceedings of the IEEE/CVF Conference on Computer Vision and Pattern Recognition (CVPR)},
  year={2020}
}

@inproceedings{liang2024skilldiffuser,
  title={Skilldiffuser: Interpretable hierarchical planning via skill abstractions in diffusion-based task execution},
  author={Liang, Zhixuan and Mu, Yao and Ma, Hengbo and Tomizuka, Masayoshi and Ding, Mingyu and Luo, Ping},
  booktitle={Proceedings of the IEEE/CVF Conference on Computer Vision and Pattern Recognition},
  pages={16467--16476},
  year={2024}
}

@article{mu2025robotwin,
  title={RoboTwin: Dual-Arm Robot Benchmark with Generative Digital Twins},
  author={Mu, Yao and Chen, Tianxing and Chen, Zanxin and Peng, Shijia and Lan, Zhiqian and Gao, Zeyu and Liang, Zhixuan and Yu, Qiaojun and Zou, Yude and Xu, Mingkun and others},
  journal={Proceedings of the IEEE/CVF conference on computer vision and pattern recognition},
  year={2025}
}

@article{liu2023libero,
  title={Libero: Benchmarking knowledge transfer for lifelong robot learning},
  author={Liu, Bo and Zhu, Yifeng and Gao, Chongkai and Feng, Yihao and Liu, Qiang and Zhu, Yuke and Stone, Peter},
  journal={Advances in Neural Information Processing Systems},
  volume={36},
  pages={44776--44791},
  year={2023}
}

@article{brohan2023rt-2,
  title={Rt-2: Vision-language-action models transfer web knowledge to robotic control},
  author={Brohan, Anthony and Brown, Noah and Carbajal, Justice and Chebotar, Yevgen and Chen, Xi and Choromanski, Krzysztof and Ding, Tianli and Driess, Danny and Dubey, Avinava and Finn, Chelsea and others},
  journal={arXiv preprint arXiv:2307.15818},
  year={2023}
}

@article{brohan2022rt-1,
  title={Rt-1: Robotics transformer for real-world control at scale},
  author={Brohan, Anthony and Brown, Noah and Carbajal, Justice and Chebotar, Yevgen and Dabis, Joseph and Finn, Chelsea and Gopalakrishnan, Keerthana and Hausman, Karol and Herzog, Alex and Hsu, Jasmine and others},
  journal={arXiv preprint arXiv:2212.06817},
  year={2022}
}

@inproceedings{openxembodiedment,
  title={Open x-embodiment: Robotic learning datasets and rt-x models: Open x-embodiment collaboration 0},
  author={O’Neill, Abby and Rehman, Abdul and Maddukuri, Abhiram and Gupta, Abhishek and Padalkar, Abhishek and Lee, Abraham and Pooley, Acorn and Gupta, Agrim and Mandlekar, Ajay and Jain, Ajinkya and others},
  booktitle={2024 IEEE International Conference on Robotics and Automation (ICRA)},
  pages={6892--6903},
  year={2024},
  organization={IEEE}
}

@article{li2025labutopia,
  title={LabUtopia: High-Fidelity Simulation and Hierarchical Benchmark for Scientific Embodied Agents},
  author={Li, Rui and Hu, Zixuan and Qu, Wenxi and Zhang, Jinouwen and Yin, Zhenfei and Zhang, Sha and Huang, Xuantuo and Wang, Hanqing and Wang, Tai and Pang, Jiangmiao and others},
  journal={arXiv preprint arXiv:2505.22634},
  year={2025}
}

@article{openvla_oft,
  title={Fine-tuning vision-language-action models: Optimizing speed and success},
  author={Kim, Moo Jin and Finn, Chelsea and Liang, Percy},
  journal={arXiv preprint arXiv:2502.19645},
  year={2025}
}

@article{zhao2022dualafford,
  title={Dualafford: Learning collaborative visual affordance for dual-gripper manipulation},
  author={Zhao, Yan and Wu, Ruihai and Chen, Zhehuan and Zhang, Yourong and Fan, Qingnan and Mo, Kaichun and Dong, Hao},
  journal={arXiv preprint arXiv:2207.01971},
  year={2022}
}

@article{liu2022robot,
  title={Robot cooking with stir-fry: Bimanual non-prehensile manipulation of semi-fluid objects},
  author={Liu, Junjia and Chen, Yiting and Dong, Zhipeng and Wang, Shixiong and Calinon, Sylvain and Li, Miao and Chen, Fei},
  journal={IEEE Robotics and Automation Letters},
  volume={7},
  number={2},
  pages={5159--5166},
  year={2022},
  publisher={IEEE}
}

@inproceedings{zakka2023robopianist,
  title={RoboPianist: Dexterous Piano Playing with Deep Reinforcement Learning},
  author={Zakka, Kevin and Wu, Philipp and Smith, Laura and Gileadi, Nimrod and Howell, Taylor and Peng, Xue Bin and Singh, Sumeet and Tassa, Yuval and Florence, Pete and Zeng, Andy and others},
  booktitle={Conference on Robot Learning},
  pages={2975--2994},
  year={2023},
  organization={PMLR}
}

@article{smith2012dual,
  title={Dual arm manipulation—A survey},
  author={Smith, Christian and Karayiannidis, Yiannis and Nalpantidis, Lazaros and Gratal, Xavi and Qi, Peng and Dimarogonas, Dimos V and Kragic, Danica},
  journal={Robotics and Autonomous systems},
  volume={60},
  number={10},
  pages={1340--1353},
  year={2012},
  publisher={Elsevier}
}

@inproceedings{lapa,
  title={Latent Action Pretraining From Videos},
  author={Ye, Seonghyeon and Jang, Joel and Jeon, Byeongguk and Joo, Se June and Yang, Jianwei and Peng, Baolin and Mandlekar, Ajay and Tan, Reuben and Chao, Yu-Wei and Lin, Bill Yuchen and others},
  booktitle={CoRL 2024 Workshop on Whole-body Control and Bimanual Manipulation: Applications in Humanoids and Beyond}
}

@inproceedings{makoviychuk2021isaac,
  title={Isaac Gym: High Performance GPU Based Physics Simulation For Robot Learning},
  author={Makoviychuk, Viktor and Wawrzyniak, Lukasz and Guo, Yunrong and Lu, Michelle and Storey, Kier and Macklin, Miles and Hoeller, David and Rudin, Nikita and Allshire, Arthur and Handa, Ankur and others},
  booktitle={NeurIPS Datasets and Benchmarks},
  year={2021}
}

@article{team2024octo,
  title={Octo: An open-source generalist robot policy},
  author={Team, Octo Model and Ghosh, Dibya and Walke, Homer and Pertsch, Karl and Black, Kevin and Mees, Oier and Dasari, Sudeep and Hejna, Joey and Kreiman, Tobias and Xu, Charles and others},
  journal={arXiv preprint arXiv:2405.12213},
  year={2024}
}

@inproceedings{wen2025dexvla,
  title={DexVLA: Vision-Language Model with Plug-In Diffusion Expert for General Robot Control},
  author={Wen, Junjie and Zhu, Yichen and Li, Jinming and Tang, Zhibin and Shen, Chaomin and Feng, Feifei},
  booktitle={Conference on Robot Learning},
  pages={3094--3114},
  year={2025},
  organization={PMLR}
}

@article{li2024cogact,
  title={Cogact: A foundational vision-language-action model for synergizing cognition and action in robotic manipulation},
  author={Li, Qixiu and Liang, Yaobo and Wang, Zeyu and Luo, Lin and Chen, Xi and Liao, Mozheng and Wei, Fangyun and Deng, Yu and Xu, Sicheng and Zhang, Yizhong and others},
  journal={arXiv preprint arXiv:2411.19650},
  year={2024}
}

@inproceedings{openvla,
  title={OpenVLA: An Open-Source Vision-Language-Action Model},
  author={Kim, Moo Jin and Pertsch, Karl and Karamcheti, Siddharth and Xiao, Ted and Balakrishna, Ashwin and Nair, Suraj and Rafailov, Rafael and Foster, Ethan P and Sanketi, Pannag R and Vuong, Quan and others},
  booktitle={8th Annual Conference on Robot Learning}
}

@article{black2024pi_0,
  title={$pi\_0 $: A Vision-Language-Action Flow Model for General Robot Control},
  author={Black, Kevin and Brown, Noah and Driess, Danny and Esmail, Adnan and Equi, Michael and Finn, Chelsea and Fusai, Niccolo and Groom, Lachy and Hausman, Karol and Ichter, Brian and others},
  journal={arXiv preprint arXiv:2410.24164},
  year={2024}
}

@article{liu2024rdt1b,
  title={Rdt-1b: a diffusion foundation model for bimanual manipulation},
  author={Liu, Songming and Wu, Lingxuan and Li, Bangguo and Tan, Hengkai and Chen, Huayu and Wang, Zhengyi and Xu, Ke and Su, Hang and Zhu, Jun},
  journal={arXiv preprint arXiv:2410.07864},
  year={2024}
}

@article{wu2024robomind,
  title={Robomind: Benchmark on multi-embodiment intelligence normative data for robot manipulation},
  author={Wu, Kun and Hou, Chengkai and Liu, Jiaming and Che, Zhengping and Ju, Xiaozhu and Yang, Zhuqin and Li, Meng and Zhao, Yinuo and Xu, Zhiyuan and Yang, Guang and others},
  journal={arXiv preprint arXiv:2412.13877},
  year={2024}
}

@article{geng2025roboverse,
  title={RoboVerse: Towards a Unified Platform, Dataset and Benchmark for Scalable and Generalizable Robot Learning},
  author={Geng, Haoran and Wang, Feishi and Wei, Songlin and Li, Yuyang and Wang, Bangjun and An, Boshi and Cheng, Charlie Tianyue and Lou, Haozhe and Li, Peihao and Wang, Yen-Jen and others},
  journal={arXiv preprint arXiv:2504.18904},
  year={2025}
}

@article{bu2025agibot,
  title={Agibot world colosseo: A large-scale manipulation platform for scalable and intelligent embodied systems},
  author={Bu, Qingwen and Cai, Jisong and Chen, Li and Cui, Xiuqi and Ding, Yan and Feng, Siyuan and Gao, Shenyuan and He, Xindong and Huang, Xu and Jiang, Shu and others},
  journal={arXiv preprint arXiv:2503.06669},
  year={2025}
}

@ARTICLE{TinyVLA,
  author={Wen, Junjie and Zhu, Yichen and Li, Jinming and Zhu, Minjie and Tang, Zhibin and Wu, Kun and Xu, Zhiyuan and Liu, Ning and Cheng, Ran and Shen, Chaomin and Peng, Yaxin and Feng, Feifei and Tang, Jian},
  journal={IEEE Robotics and Automation Letters}, 
  title={TinyVLA: Toward Fast, Data-Efficient Vision-Language-Action Models for Robotic Manipulation}, 
  year={2025},
  volume={10},
  number={4},
  pages={3988-3995},
  doi={10.1109/LRA.2025.3544909}}

@inproceedings{grotz2024peract2,
  title={Peract2: Benchmarking and learning for robotic bimanual manipulation tasks},
  author={Grotz, Markus and Shridhar, Mohit and Chao, Yu-Wei and Asfour, Tamim and Fox, Dieter},
  booktitle={CoRL 2024 Workshop on Whole-body Control and Bimanual Manipulation: Applications in Humanoids and Beyond},
  year={2024}
}

@article{chen2025robotwin,
  title={Robotwin 2.0: A scalable data generator and benchmark with strong domain randomization for robust bimanual robotic manipulation},
  author={Chen, Tianxing and Chen, Zanxin and Chen, Baijun and Cai, Zijian and Liu, Yibin and Li, Zixuan and Liang, Qiwei and Lin, Xianliang and Ge, Yiheng and Gu, Zhenyu and others},
  journal={arXiv preprint arXiv:2506.18088},
  year={2025}
}

@article{lan2025autobio,
  title={Autobio: A simulation and benchmark for robotic automation in digital biology laboratory},
  author={Lan, Zhiqian and Jiang, Yuxuan and Wang, Ruiqi and Xie, Xuanbing and Zhang, Rongkui and Zhu, Yicheng and Li, Peihang and Yang, Tianshuo and Chen, Tianxing and Gao, Haoyu and others},
  journal={arXiv preprint arXiv:2505.14030},
  year={2025}
}

@inproceedings{metaworld,
  title={Meta-world: A benchmark and evaluation for multi-task and meta reinforcement learning},
  author={Yu, Tianhe and Quillen, Deirdre and He, Zhanpeng and Julian, Ryan and Hausman, Karol and Finn, Chelsea and Levine, Sergey},
  booktitle={Conference on robot learning},
  pages={1094--1100},
  year={2020},
  organization={PMLR}
}

@inproceedings{li2023efficient,
  title={Efficient bimanual handover and rearrangement via symmetry-aware actor-critic learning},
  author={Li, Yunfei and Pan, Chaoyi and Xu, Huazhe and Wang, Xiaolong and Wu, Yi},
  booktitle={2023 IEEE International Conference on Robotics and Automation (ICRA)},
  pages={3867--3874},
  year={2023},
  organization={IEEE}
}

@inproceedings{grannen2023stabilize,
  title={Stabilize to act: Learning to coordinate for bimanual manipulation},
  author={Grannen, Jennifer and Wu, Yilin and Vu, Brandon and Sadigh, Dorsa},
  booktitle={Conference on Robot Learning},
  pages={563--576},
  year={2023},
  organization={PMLR}
}

@inproceedings{wang2024rise,
  title={Rise: 3d perception makes real-world robot imitation simple and effective},
  author={Wang, Chenxi and Fang, Hongjie and Fang, Hao-Shu and Lu, Cewu},
  booktitle={2024 IEEE/RSJ International Conference on Intelligent Robots and Systems (IROS)},
  pages={2870--2877},
  year={2024},
  organization={IEEE}
}

@inproceedings{chen2025g3flow,
  title={G3flow: Generative 3d semantic flow for pose-aware and generalizable object manipulation},
  author={Chen, Tianxing and Mu, Yao and Liang, Zhixuan and Chen, Zanxin and Peng, Shijia and Chen, Qiangyu and Xu, Mingkun and Hu, Ruizhen and Zhang, Hongyuan and Li, Xuelong and others},
  booktitle={Proceedings of the Computer Vision and Pattern Recognition Conference},
  pages={1735--1744},
  year={2025}
}

@inproceedings{dexhanddiff,
  title={DexHandDiff: Interaction-aware Diffusion Planning for Adaptive Dexterous Manipulation},
  author={Liang, Zhixuan and Mu, Yao and Wang, Yixiao and Chen, Tianxing and Shao, Wenqi and Zhan, Wei and Tomizuka, Masayoshi and Luo, Ping and Ding, Mingyu},
  booktitle={Proceedings of the Computer Vision and Pattern Recognition Conference},
  pages={1745--1755},
  year={2025}
}

@article{chi2023diffusion,
  title={Diffusion policy: Visuomotor policy learning via action diffusion},
  author={Chi, Cheng and Xu, Zhenjia and Feng, Siyuan and Cousineau, Eric and Du, Yilun and Burchfiel, Benjamin and Tedrake, Russ and Song, Shuran},
  journal={The International Journal of Robotics Research},
  pages={02783649241273668},
  year={2023},
  publisher={SAGE Publications Sage UK: London, England}
}

@inproceedings{ke20243d,
  title={3D Diffuser Actor: Policy Diffusion with 3D Scene Representations},
  author={Ke, Tsung-Wei and Gkanatsios, Nikolaos and Fragkiadaki, Katerina},
  booktitle={Conference on Robot Learning},
  pages={1949--1974},
  year={2025},
  organization={PMLR}
}

@inproceedings{robocasa2024,
    title={RoboCasa: Large-Scale Simulation of Everyday Tasks for Generalist Robots},
    author={Soroush Nasiriany and Abhiram Maddukuri and Lance Zhang and Adeet Parikh and Aaron Lo and Abhishek Joshi and Ajay Mandlekar and Yuke Zhu},
    booktitle={Robotics: Science and Systems (RSS)},
    year={2024}
}

@inproceedings{lv2025spatial,
  title={Spatial-temporal graph diffusion policy with kinematic modeling for bimanual robotic manipulation},
  author={Lv, Qi and Li, Hao and Deng, Xiang and Shao, Rui and Li, Yinchuan and Hao, Jianye and Gao, Longxiang and Wang, Michael Yu and Nie, Liqiang},
  booktitle={Proceedings of the Computer Vision and Pattern Recognition Conference},
  pages={17394--17404},
  year={2025}
}

@article{yang2025gripper,
  title={Gripper Keypose and Object Pointflow as Interfaces for Bimanual Robotic Manipulation},
  author={Yang, Yuyin and Cai, Zetao and Tian, Yang and Zeng, Jia and Pang, Jiangmiao},
  journal={arXiv preprint arXiv:2504.17784},
  year={2025}
}

@inproceedings{yu2025manigaussian++,
  title={ManiGaussian++: General robotic bimanual manipulation with hierarchical Gaussian world model},
  author={Yu, Tengbo and Lu, Guanxing and Yang, Zaijia and Deng, Haoyuan and Chen, Season Si and Lu, Jiwen and Ding, Wenbo and Hu, Guoqiang and Tang, Yansong and Wang, Ziwei},
  booktitle={2025 IEEE/RSJ International Conference on Intelligent Robots and Systems (IROS)},
  pages={12232--12239},
  year={2025},
  organization={IEEE}
}

@article{hao2026abstracting,
  title={Abstracting Robot Manipulation Skills via Mixture-of-Experts Diffusion Policies},
  author={Hao, Ce and Zhai, Xuanran and Liu, Yaohua and Soh, Harold},
  journal={arXiv preprint arXiv:2601.21251},
  year={2026}
}

@inproceedings{liu2025d,
  title={D-CODA: Diffusion for Coordinated Dual-Arm Data Augmentation},
  author={Liu, I-Chun Arthur and Chen, Jason and Sukhatme, Gaurav S and Seita, Daniel},
  booktitle={Conference on Robot Learning},
  pages={3569--3588},
  year={2025},
  organization={PMLR}
}

@inproceedings{jiang2025rethinking,
  title={Rethinking bimanual robotic manipulation: Learning with decoupled interaction framework},
  author={Jiang, Jian-Jian and Wu, Xiao-Ming and He, Yi-Xiang and Zeng, Ling-An and Wei, Yi-Lin and Zhang, Dandan and Zheng, Wei-Shi},
  booktitle={Proceedings of the IEEE/CVF International Conference on Computer Vision},
  pages={12427--12437},
  year={2025}
}

@inproceedings{liu2025voxact,
  title={VoxAct-B: Voxel-Based Acting and Stabilizing Policy for Bimanual Manipulation},
  author={Liu, I-Chun Arthur and He, Sicheng and Seita, Daniel and Sukhatme, Gaurav S},
  booktitle={Conference on Robot Learning},
  pages={4354--4370},
  year={2025},
  organization={PMLR}
}

@inproceedings{gao2024bi,
  title={Bi-kvil: Keypoints-based visual imitation learning of bimanual manipulation tasks},
  author={Gao, Jianfeng and Jin, Xiaoshu and Krebs, Franziska and Jaquier, No{\'e}mie and Asfour, Tamim},
  booktitle={2024 IEEE International Conference on Robotics and Automation (ICRA)},
  pages={16850--16857},
  year={2024},
  organization={IEEE}
}

@inproceedings{lu2025anybimanual,
  title={Anybimanual: Transferring unimanual policy for general bimanual manipulation},
  author={Lu, Guanxing and Yu, Tengbo and Deng, Haoyuan and Chen, Season Si and Tang, Yansong and Wang, Ziwei},
  booktitle={Proceedings of the IEEE/CVF International Conference on Computer Vision},
  pages={13662--13672},
  year={2025}
}

\end{document}